\documentclass{article}

\usepackage{graphicx} 
\usepackage{subfigure}
\usepackage{xspace}
\usepackage{color}

\usepackage{natbib}

\usepackage{algorithm}
\usepackage{algorithmic}

\usepackage{amsmath,enumerate}

\usepackage{hyperref}

\newcommand{\comment}[1]{}
\renewcommand{\cite}[1]{\citep{#1}}

\usepackage{color}
\newtheorem{theorem}{Theorem} 

\newcommand{\Tr}{^{\rm T}}

\newcommand{\s}{{\bf s}}

\renewcommand{\v}{{\bf v}}
\newcommand{\w}{{\bf w}}
\newcommand{\x}{{\bf x}}
\newcommand{\y}{{\bf y}}

\newcommand{\A}{{\bf A}}

\newcommand{\N}{{\cal N}}  

\newcommand{\V}{{\bf V}}

\newcommand{\X}{{\bf X}}

\newcommand{\balpha}{\boldsymbol{\alpha}}

\newcommand{\bbeta}{\boldsymbol{\beta}}

\newcommand{\0}{{\bf 0}}

\newcommand{\ben}{\begin{enumerate}}
\newcommand{\een}{\end{enumerate}}

\newcommand{\dif}{\mathrm{d}}

\begin{document}

\newcommand{\eig}{EigenNet\xspace}
\newcommand{\fengc}[1]{[\textcolor{blue}{Feng writes: #1}]}
\newcommand{\alanc}[1]{[\textcolor{red}{Alan writes: #1}]}
\newcommand{\tw}{\tilde{\w}}

\title{
EigenNet: A Bayesian hybrid of generative and conditional models for sparse learning \\
}

\author{Yuan Qi \\
        Departments of CS and Statistics \\
        Purdue University
        \and
        Feng Yan \\
        Department of CS\\ Purdue University \\
}

\maketitle



\begin{abstract}
It is a challenging task to select correlated variables in a high dimensional space. To address this challenge,
the elastic net has been developed and successfully applied to many applications. Despite its great success, the elastic
net does not explicitly use correlation information embedded in data to select correlated variables. To overcome this limitation,
we present a novel Bayesian hybrid model, the \eig, that uses the eigenstructures of data to guide variable selection. Specifically,
it integrates a sparse conditional classification model with a generative model capturing variable correlations in a principled Bayesian framework.
We reparameterize the hybrid model in the eigenspace to avoid overfiting and to increase the computational efficiency of its MCMC sampler.
Furthermore, we provide an alternative view to the \eig from a regularization perspective: the \eig has an adaptive eigenspace-based composite regularizer, which naturally generalizes the $l_{1/2}$ regularizer used by the elastic net. Experiments on synthetic and real data show that the \eig significantly outperforms the lasso, the elastic net, and the Bayesian lasso in terms of prediction accuracy, especially when the number of training samples is smaller than the number of variables.
\end{abstract}

\section{Introduction}
In this paper we consider the problem of selecting correlated variables in a high dimensional space.
Among many variable selection methods, the lasso and the elastic net are two popular choices \cite{Tibshirani94lasso,ZouHastie05elasticnet}.
The lasso uses a $l_1$ regularizer on model parameters. This regularizer shrinks the parameters towards zero, removing irreverent variables and yielding a sparse model~\cite{Tibshirani94lasso}. However, the $l_1$ penalty may lead to over-sparisification: given many correlated variables, the lasso often only select a few of them.
This not only degenerates its prediction accuracy but also affects the interpretability of the estimated model.
For example, based on high-throughput biological data such as gene expression and RNA-seq data, it is highly desirable to select multiple correlated genes specific to a phenotype since it may reveal underlying biological pathways.  Due to its over-sparsification, lasso may not be suitable for this task.


To address this issue, the elastic net has been developed to encourage a grouping
effect, where strongly correlated variables tend to be in or out of the model together~\cite{ZouHastie05elasticnet}.
However, the grouping effect is just the result of its composite $l_1$ and $l_2$ regularizer; the elastic net does not explicitly incorporate correlation information among variables in its model. 

In this paper, we propose a new sparse Bayesian hybrid model, called  the {\em \eig}. Unlike the previous sparse models, it uses the eigen information from the data covariance matrix to guide the selection of correlated variables. Specifically, it integrates a sparse {\em conditional} classification model with a {\em generative} model capturing variable correlation in a principle Bayesian framework~\cite{Lass06}.
The hybrid model enables identification of groups of correlated variables guided by the eigenstructures. 
 Also, it passes the information from the conditional model to the generative model, selecting informative eigenvectors for the classification task. Unlike frequentist approaches, the Bayesian hybrid model can 
 reveal correlations between classifier weights via their joint posterior distribution.

We reparameterize the model in the eigenspace of the data. When the number of predictor variables (i.e., input features), $(p)$, is bigger than the number of training samples $(n)$, this reparameterization restricts the model in the data subspace, which not only reduces overfitting, but also allows us to develop efficient Markov Chain Monte Carlo sampler.

From the regularization perspective, the \eig  naturally generalizes the elastic net by using a composite regularizer adaptive to the data eigenstructures. It contains a $l_1$ sparsity regularizer and a directional regularizer that encourages selecting variables associated with eigenvectors chosen by the model. When the variables are independent of each other, the eigenvectors are parallel to the axes and this composite regularizer reduces to the $l_{1/2}$ regularizer used by the elastic net; when some of the input variables are strongly correlated, the regularizer will encourage the classifier aligned with eigenvectors selected by the model. 
On one hand, our model is like the elastic net to retain `all the big fish'.  On the other hand, our model is different from the elastic net by using the eigenstructure. Hence the name \eig.

Experiments on synthetic and real data are presented in Section 7. They demonstrate that the \eig significantly outperforms the lasso, the elastic net, and the Bayesian lasso \cite{ParTreCasGeo08BayLasso,Hans09BayLasso} in terms of prediction accuracy, especially when the number of training samples is smaller than the number of features.

\section{Background: lasso and elastic net}

We denote $n$ independent and identically distributed samples as
$$\mathcal{D}=\{(\x_{1},y_{1}),\ldots,(\x_{n},y_{n})\}$$, where $\x_i$ is a $p$ dimensional input features (i.e., explanatory variables) and $y_i$ is a scalar label (i.e., response). Also, we denote $[\x_1,\ldots,\x_n]$ by $\X$ and $(\y_1,\ldots,\y_n)$ by $\y$.
In this paper, we consider the binary classification problem ($y_i\in \{-1,1\}$), but our analysis and the proposed models can be extended to regression and other problems.

For classification, we use a logistic function as the data likelihood function:
\begin{align} \label{eq:lik_w}
p(\y|\X, \w, b) &= \prod_i \sigma(y_i (\w\Tr \x_i + b) )
\end{align}
where $\sigma(z) = \frac {1} {1+\rm{exp}(-z)}$, and $\w$ and $b$ define the classifier.

To identify relevant variables for high dimensional problems, the lasso \cite{Tibshirani94lasso}
uses a $l_1$ penalty, effectively shrinking $\w $ and $b$ towards zero and pruning irrelevant variables.
In a probabilistic framework this penalty corresponds to a Laplace prior distribution:
\begin{align}
p(\w) &  =  \prod_j \lambda \exp(-\lambda |w_j|)
\end{align}
where $\lambda$ is a hyperparameter that controls the sparsity of the estimated model. The larger the hyperparameter $\lambda$, the sparser the model.

As described in Section 1, the lasso may over-penalize relevant variables and hurt its predictive performance, especially when there are strongly correlated variables.
To address this issue, the elastic net \cite{ZouHastie05elasticnet} combines $l_1$ and $l_2$ regularizers to avoid the over-penalization. The combined regularizer corresponds to the following prior distribution:
\begin{align}
p(\w) & \propto  \prod_j \exp (-\lambda_1 |w_j| - \lambda_2 w_j^2 )
\end{align}
where $\lambda_1$ and $\lambda_2$ are hyperparameters.
While it is well known that the elastic net tends to select strongly correlated variables together, it does not uses correlation information embedded in the data. The selection of correlated variables is merely the result of a less aggressive regularizer for sparisty.

Besides the elastic net, there are many variants (and extensions) to the lasso, such as the bridge \cite{Frank93} and smoothly clipped absolute deviation \cite{FanLi01SCAD}. These variants modify the $l_1$ penalty to choose variables, but again do not explicitly use correlation information in data.

\section{EigenNet: eigenstructure-guided variable selection}

In this section, we propose to use covariance structures in data to guide the sparse estimation of model parameters.

First, let us consider the following toy examples.

\subsection{Toy examples}
\begin{figure}[h]
\vspace{-0.1in}
\center
{
\subfigure[Independent variables]{\label{fig:ind_vars}\includegraphics[width=2.1in]{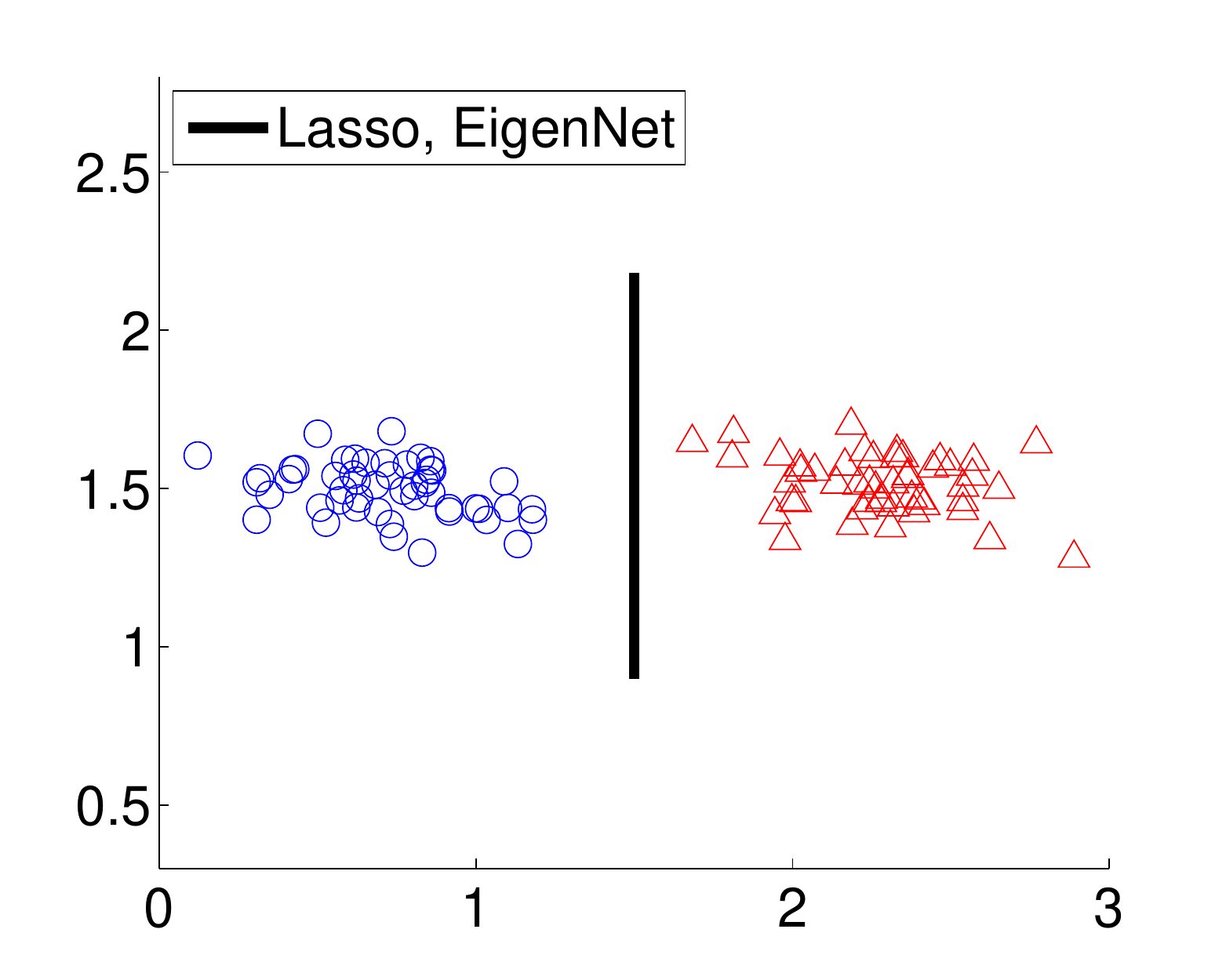}}
\subfigure[Correlated variables]{\label{fig:cor_vars}\includegraphics[width=2.1in]{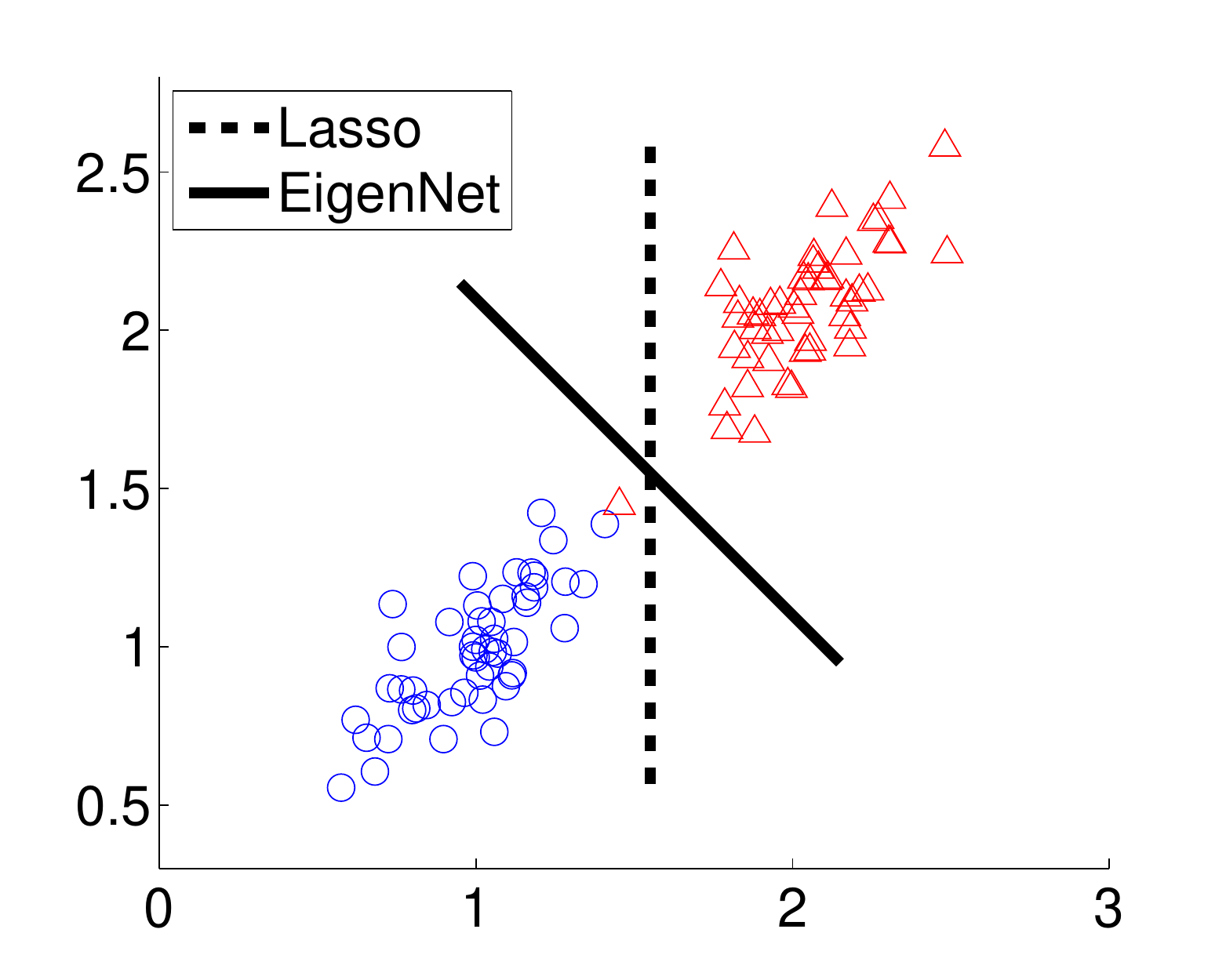}}
}
\vspace{-10pt}
\caption{Toy examples. (a) When the variables $x^1$ and $x^2$ are independent of each other, both the lasso and the \eig select only $x^1$. (b) When the variables $x^1$ and $x^2$ are correlated, the lasso selects only one variable. By contrast, guided by the major eigenvector of the data, the \eig selects both variables.}
\end{figure}

Figure \ref{fig:ind_vars} shows samples from two classes. Clearly the variables $x^1$ and $x^2$ are not correlated. The lasso or the elastic net can successfully select the relevant variable $x^1$ to classify the data. For the samples in Figure \ref{fig:cor_vars},  the variables $x^1$ and $x^2$ are  strongly correlated. Despite the strong correlation, the lasso would select only $x^1$ and ignore $x^2$. The elastic net may select both $x^1$ and $x^2$ if the regularization weight $\lambda_1$ is small and $\lambda_2$ is big, so that the elastic net behaves like $l_2$ regularized classifier. The elastic net, however, does not explore the fact that $x^1$ and $x^2$ are correlated.

%

Since the eigenstructure of the data covariance matrix captures correlation information between variables, we propose
to not only regularize the classifier to be sparse, but also encourage it to be aligned with certain eigenvector(s) that are helpful for the classification task.
Since our new model uses the eigen information, we name it the \eig.

For the data in Figure \ref{fig:ind_vars}, since the two eigenvectors are parallel with the horizontal and vertical axes, the \eig essentially reduces to the elastic net and selects $x^1$. For the data in Figure \ref{fig:cor_vars}, however, the eigenvectors (in particular, the principle eigenvector) will guide the \eig to select both $x^1$ and  $x^2$.

We use a Bayesian framework to materialize the above ideas in the \eig, as shown in the following section.

\subsection{Bayesian hybrid of conditional and generative models}

The \eig is a hybrid of conditional and generative models. The conditional component allows us to learn the classifier via "discriminative" training; the generative component captures the correlations between variables; and these two models are glued together via a joint prior distribution, so that the correlation information is used to guide the estimation of the classifier and the classification task is used to choose or scale relevant eigenvectors.
Our approach is based on the general Bayesian framework proposed by \citet{Lass06}), which allows one to combine
conditional and generative models in an elegant principled way.

Specifically, for the conditional model we have the same likelihood as \eqref{eq:lik_w}, 
$p(\y|\X,\w,b)  = \prod_i \sigma(y_i (\w\Tr \x_i + b))$.
To sparsify the classifier, we can use a Laplace prior on $\w$,
\begin{align}
p(\w) &  =  \prod_j \lambda_1 \exp\{-\lambda_1 |w_j|\}.
\end{align}

To encourage the classifier aligned with certain eigenvectors, we use the following generative model:
\begin{align}
p(\V \s| \tw)  & \propto \exp( -\frac{\lambda_2}{2} \sum \eta_j || \tw - s_i \v_i ||^2_+)
\end{align}
where
\begin{align} \label{eq:wv}
& || \tw - s_i \v_i ||^2_+  \nonumber \\
\equiv & - \frac{1}{2} \lambda_2 \sum_j \eta_j ( ||\tw||^2 - 2 s_j |\tw\Tr \v_j| + s_j^2 ||\v_j||^2 \nonumber \\
= & - \frac{1}{2} \lambda_2 \sum_j \eta_j ( ||\tw||^2 - 2 s_j |\tw\Tr \v_j| + s_j^2,
\end{align}
 $\s$ are nonnegative continuous variables,
 $\v_i$ and $\eta_i$ are the $i$-th eigenvector and eigenvalue of the data covariance matrix, respectively.
The reason we use absolute values of $\tw\Tr\v_j$ in \eqref{eq:wv} is because we only care about the alignment of $\tw$ and $\v_i$, not the sign of their product. Overall, the above model encourages the classifier to more aligned with the major eigenvectors with bigger eigenvalues. But the variables $\s$ allow us to scale or select individual eigenvectors to remove irrelevant ones.

\begin{figure}[t]
\center
\label{fig:abaloneicf}
\includegraphics[scale=0.5]{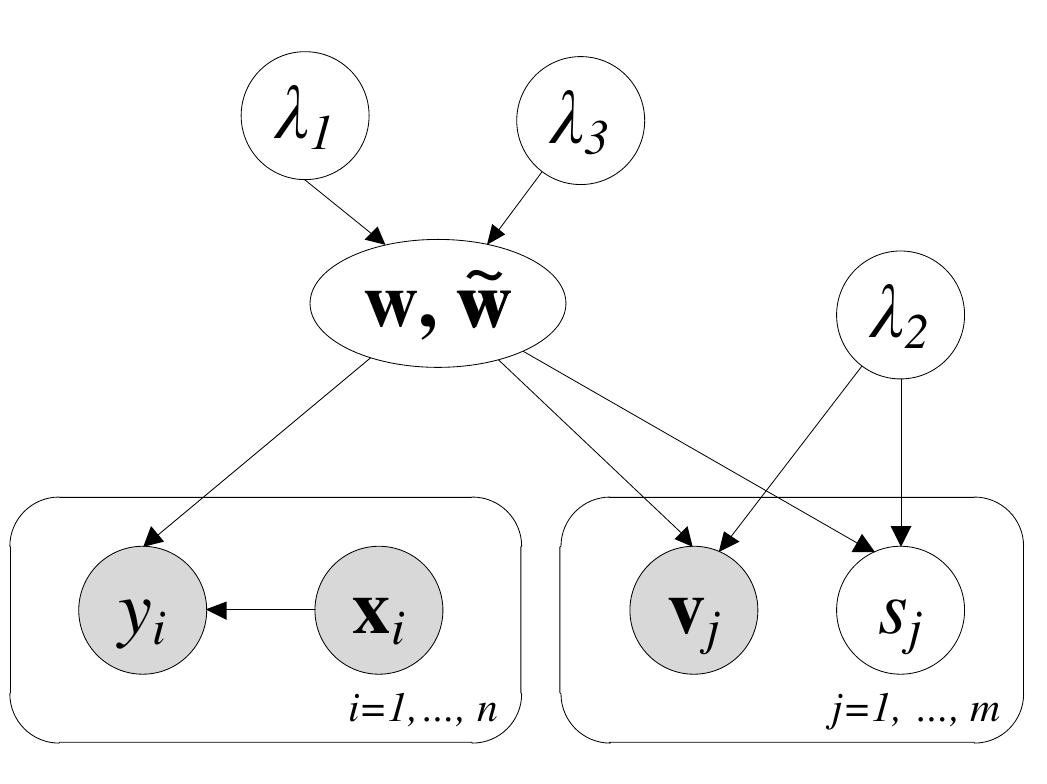}
\vspace{-7pt}
\caption{The graphical model of the \eig.}
\label{fig:joint_model}
\end{figure}

To integrate the conditional and generative models, we use a joint prior on $\w$ and $\tw$:
\begin{align}
p(\w,\tw) \propto  \exp(-\lambda_1 |\w|_1) \exp(-\frac{\lambda_3}{2} ||\w - \tw ||^2).
\end{align}
i.e., we have
\begin{align} \label{eq:prior_joint}
p(\w,\tw) = \lambda_1 \exp(-\lambda_1 |\w|_1) \N(\tw|\w, {\lambda_3}^{-1}).
\end{align}
Finally we can assign Gamma priors on all the hyperparameters, $\lambda_1$, $\lambda_2$, and $\lambda_3$. The whole model is depicted in the graphical model in Figure \ref{fig:joint_model}.


\subsection{Reparameterization and constraint in Eigenspace}

In this section we reparameterize the model in the eigenspace:
\begin{align}\label{eq:wa}
\w= \V \balpha & \quad \tw= \V \bbeta
\end{align}
where $\V \equiv [\v_1, \ldots, \v_m]$ $(m=\min\{n,p\})$, and $\balpha$ and $\bbeta$ are the projections of $\w$ and $\tw$
on the eigenvectors, respectively.

The reparameterization restricts $\w$ in the vector space spanned by $\{\v_1,\ldots,\v_m\}$, which is equivalent to
the data space $\cal{C}(\X)$, spanned by the data points $\{\x_1,\ldots,\x_n\}$.
When the number of features is bigger than the number of training points, {\em i.e.}, $p>n$, it effectively reduces
 the number of free parameters in the model, helping avoid overfitting. Furthermore,
it provides significant computational advantage when  $p>>n$.

Given $p(\w, \tw)$ and the relationship between $(\w, \tw)$ and $(\balpha, \bbeta)$, we obtain $p(\balpha, \bbeta)$ (Please see Appendix for the details):
\begin{align}\label{eq:prior0}
p(\balpha,\bbeta) \propto  \exp(-\lambda_1 |\V \balpha|_1) \exp(-\frac{\lambda_3}{2} ||\balpha - \bbeta ||^2)
\end{align}

Based on the new reparameterization, the likelihood for the conditional model becomes
\begin{align}\label{eq:lik_alpha}
p(\y|\X, \balpha, b) = \prod_i \sigma(y_i (\x_i\Tr \V \balpha + b) ).
\end{align}
Similarly, the likelihood for the generative model becomes
\begin{align}\label{eq:lik_beta}
p(\V, \s | \bbeta)  & \propto \exp(- \frac{1}{2}  \lambda_2 \sum_j \eta_j ( ||\V \bbeta||^2  \nonumber \\
                    & - 2 s_j |\V \bbeta\Tr \v_j| + ||\v_j||^2) \nonumber \\
                    & \propto \exp(- \frac{1}{2} \lambda_2 \sum_j ( \beta_j^2 - 2 \eta_j s_j |\beta_j| + \eta_j s_j^2 ))
\end{align}
The second equation holds since $\V$ is an orthonormal matrix.

Combining \eqref{eq:prior0}, \eqref{eq:lik_alpha} and \eqref{eq:lik_beta}, we obtain a complete model. We use Markov Chain Monte Carlo with a random walk proposal to estimate the model parameters $\s$, $\w$, and $\tw$.


\begin{figure}[tb]
\center
{
\subfigure[]{\label{fig:eignet_ind}\includegraphics[scale=0.65]{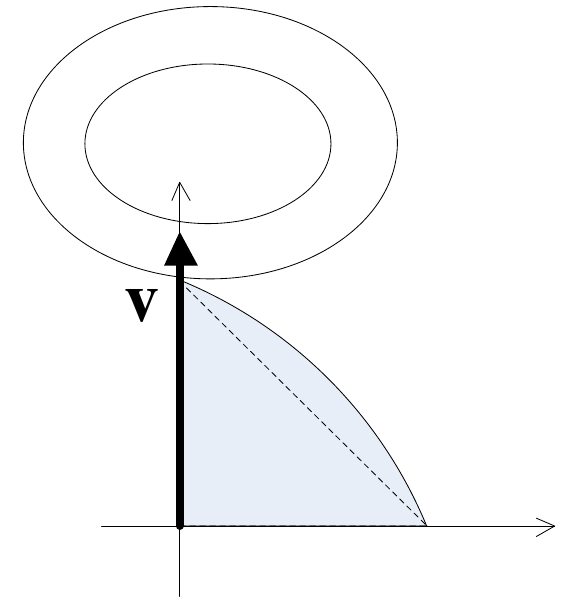}}
\subfigure[]{\label{fig:eignet_cor}\includegraphics[scale=0.65]{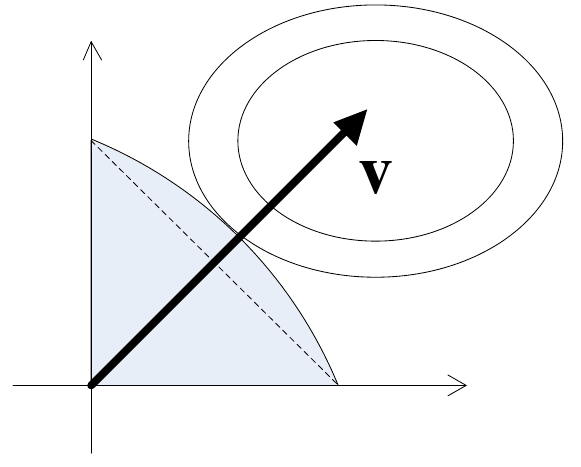}}
}
\vspace{-10pt}
\caption{Adaptive regularization of the \eig. The ellipses are the contours of a likelihood function. While the lasso draws the estimates towards the $l_1$ ball, the \eig's estimate is guided by an eigenvector $\v$.
}
\label{fig:regularizers}
\end{figure}

\section{Alternative view: composite regularization}
In this section, we provide an alternative view to the \eig by considering the limiting case of $\lambda_3\rightarrow 0$.
For such as case the prior $p(\balpha, \bbeta)$ becomes
\[
p(\balpha, \bbeta) = p(\balpha) \delta(\balpha - \bbeta)
\]
This forces $\balpha = \bbeta$. From a regularization perspective, this prior is equivalent to a composite regularizer:
\begin{align}
&\lambda_1 |\w| + \frac{\lambda_2}{2} \sum \eta_j || \w - s_j \v_j ||^2_+ \\
=& \lambda_1 |\w| + \frac{\lambda_2}{2} \sum \eta_j (||\w||^2 - 2 s_j |\w\Tr\v_j| + s_j^2)
\end{align}
Clearly, when $s_i = 0$ for all $i$'s, the above regularizer reduces to the $l_{1/2}$ regularizer used by the elastic net \footnote{A subtle difference is that we also constrain $\w$ in the data space for our model.}.
When  $s_i \neq 0$ then the regularizer is {\em adaptive} based on the eigenvector $\v_i$: First, if the elements of $\v_i$ all have reasonably large values, then all the variables in $\w$ will very likely to be selected. This effect is visualized in Figure \ref{fig:eignet_cor}. Second, if this
eigenvector has only several large elements, the corresponding variables in $\tw$ and $\w$ are likely to be selected jointly. Unlike the $l_{1/2}$ regularizer that encourages the selection of groups of variables from all the variables, our regularizer directly targets at {\em specific} groups of variables corresponding to the sparse eigenvector. Third, if all the variables are independent of each other, then the eigenvectors are parallel to the axes and each of them contains only one nonzero element. In this case $|\w\Tr\v_j|$ reduces $|w_j|$, a $l_1$ regularizer.
Figure \ref{fig:eignet_ind} visualizes the eigen regularizer when variables are independent of each other.

In summary, the \eig can be viewed as an adaptive generalization of the elastic net by selecting groups of correlated variables based on eigenvectors of the data covariance matrix.

\section{Related work}

The \eig can be viewed as an extension of the classical eigenface approaches \cite{TurkPent91,Sirovich87}. The eigenface approach uses PCA coefficients of samples to train a classifier. Naturally the major eigenvectors are often associated with large PCA coefficients and the classifier is constrained in the data subspace when the number of features is smaller than the number of training samples. The \eig essentially extends the eigenface approach by combining generative and conditional models in a Bayesian framework and performs sparse learning in an adaptive eigenspace (since the model selects or scales relevant eigenvectors based on $s_j$).

There are Bayesian versions of the lasso and the elastic net. Bayesian lasso \cite{ParTreCasGeo08BayLasso} puts a hyper-prior on the regularization coefficient and use a Gibbs sampler to jointly sample both regression weights and the regularization coefficient. Using a similar treatment to Bayesian lasso, Bayesian elastic net \cite{LiLin10BayElastic} samples the two regularization coefficients simultaneously, potentially avoiding the ``double shrinkage" problem described in the original elastic net paper \cite{ZouHastie05elasticnet}. As the \eig, these methods are grounded in a Bayesian framework, sharing the benefits of obtaining posterior distributions for handling estimation uncertainty. However, Bayesian lasso and Bayesian elastic net are presented to handle regression problems (though certainly they can be generalized for classification problems) and sample in the original parameter space, not using the eigen information embedded in data. The \eig, by contrast, works in the eigenspace and uses eigen information to guide classification.

\section{Experimental results}

We evaluate the new sparse Bayesian model, the \eig, on both synthetic and real data and compare it with three representative state-of-the-art variable selection methods, including the lasso, the elastic net, and the Bayesian lasso modified for classification problems. For the lasso and the elastic net we use the Glmnet software package that uses cyclical coordinate descent in a pathwise fashion\footnote{http://www-stat.stanford.edu/~tibs/glmnet-matlab/}. 
The original Bayesian lasso was developed for regression and uses Gibbs sampling. For the classification tasks we consider, we change its Gaussian regression likelihood to the logistic likelihood \eqref{eq:lik_w} while keeping its Laplace prior distributions. We used Markov Chain Monte Carlo, instead of Gibbs sampler, to estimate the classifier for the Bayesian lasso. Bayesian approaches are capable of estimating all the hyperparameters from data. However, for easy and objective comparisons, we simply use cross-validation to tune the hyperparameters, $\lambda_i$, for all methods. For the Bayesian lasso and the \eig, we draw the 300,000 MCMC samples and use the last 150,000 samples to estimate the posterior mean of the classifiers, which are used for predicting the labels of test samples. We measure the prediction performance of all methods on test samples in terms of their average test error rate (e.g., the 0.2 error rate indicates $20\%$ errors) and report the standard error of the error rates (except for the following visualization example).

\begin{figure}[tb]
\center
{
\subfigure[Lasso]{\label{fig:w_lasso}\includegraphics[width=1.6in]{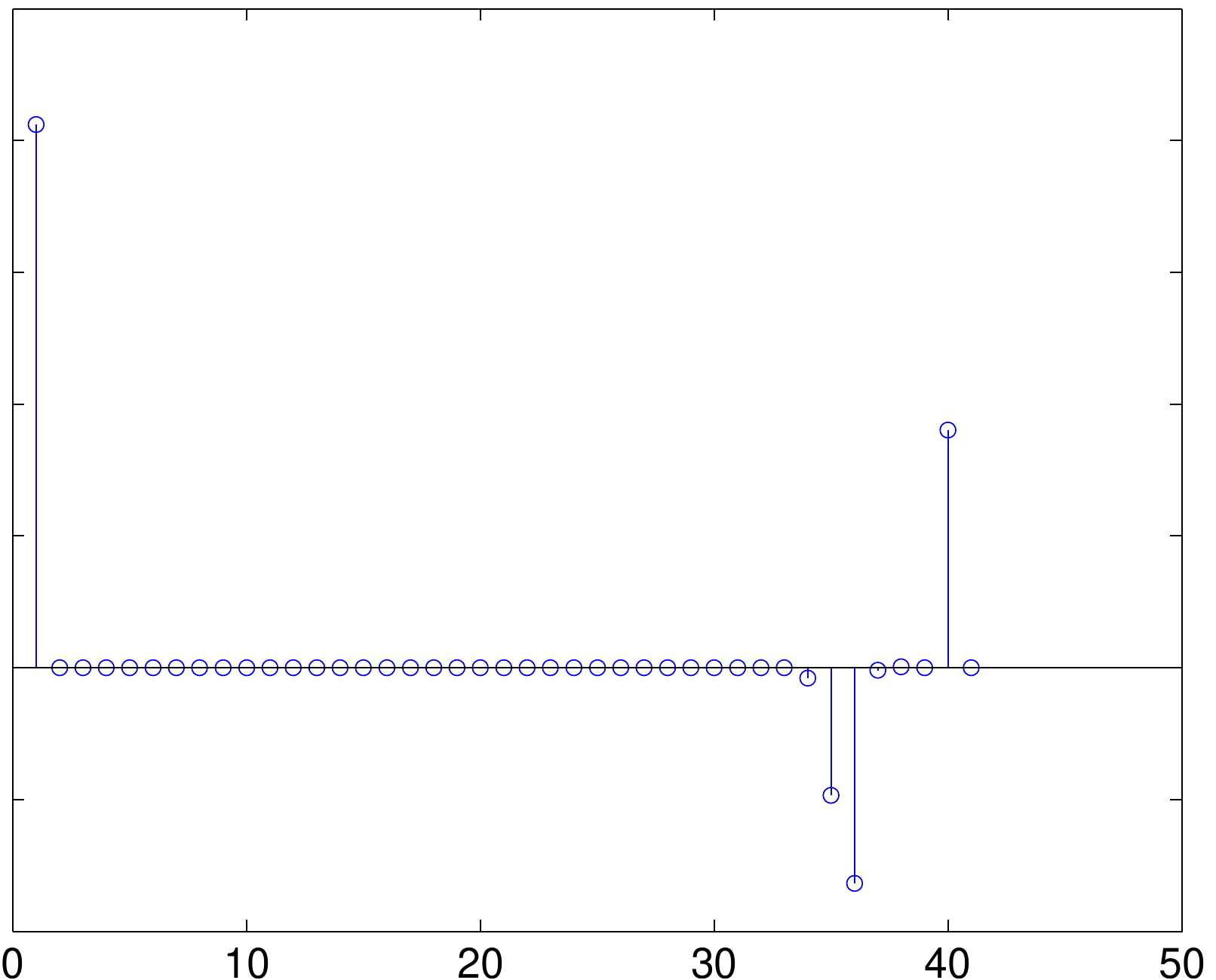}}
\subfigure[Elastic net]{\label{fig:w_elasticnet}\includegraphics[width=1.6in]{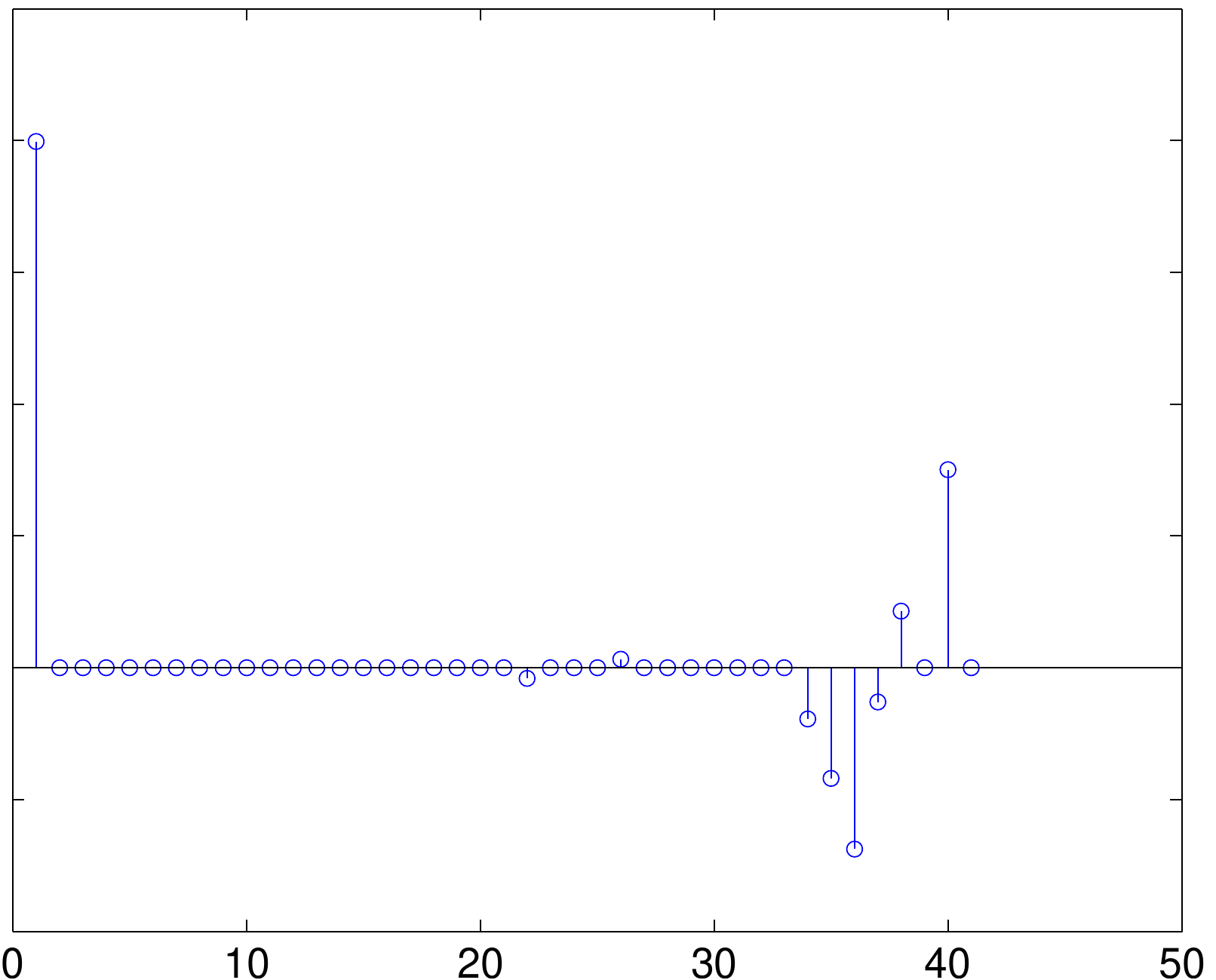}}
\subfigure[\eig]{\label{fig:w_eigennet}\includegraphics[width=1.6in]{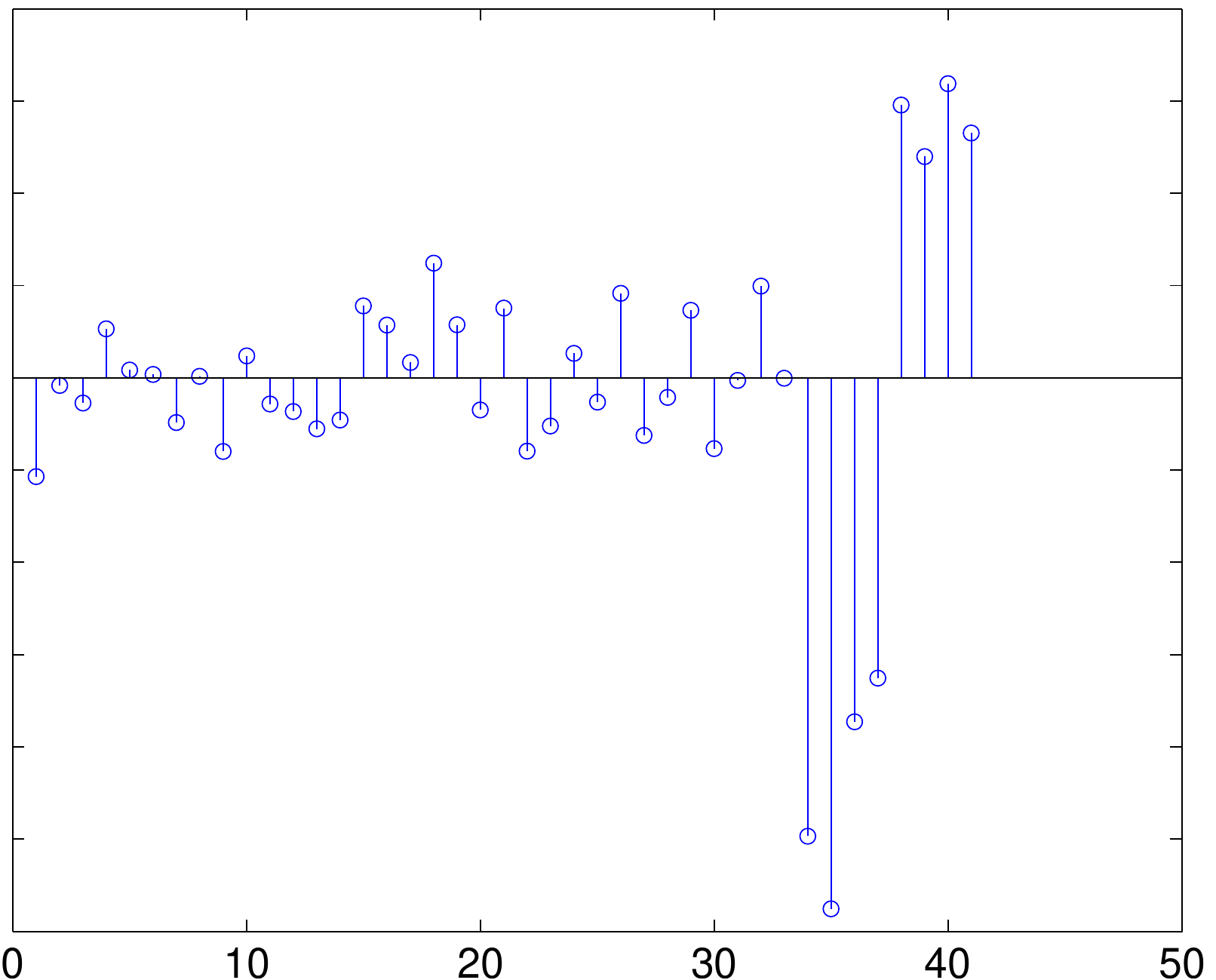}}
\subfigure[True]{\label{fig:w_w_tr}\includegraphics[width=1.6in]{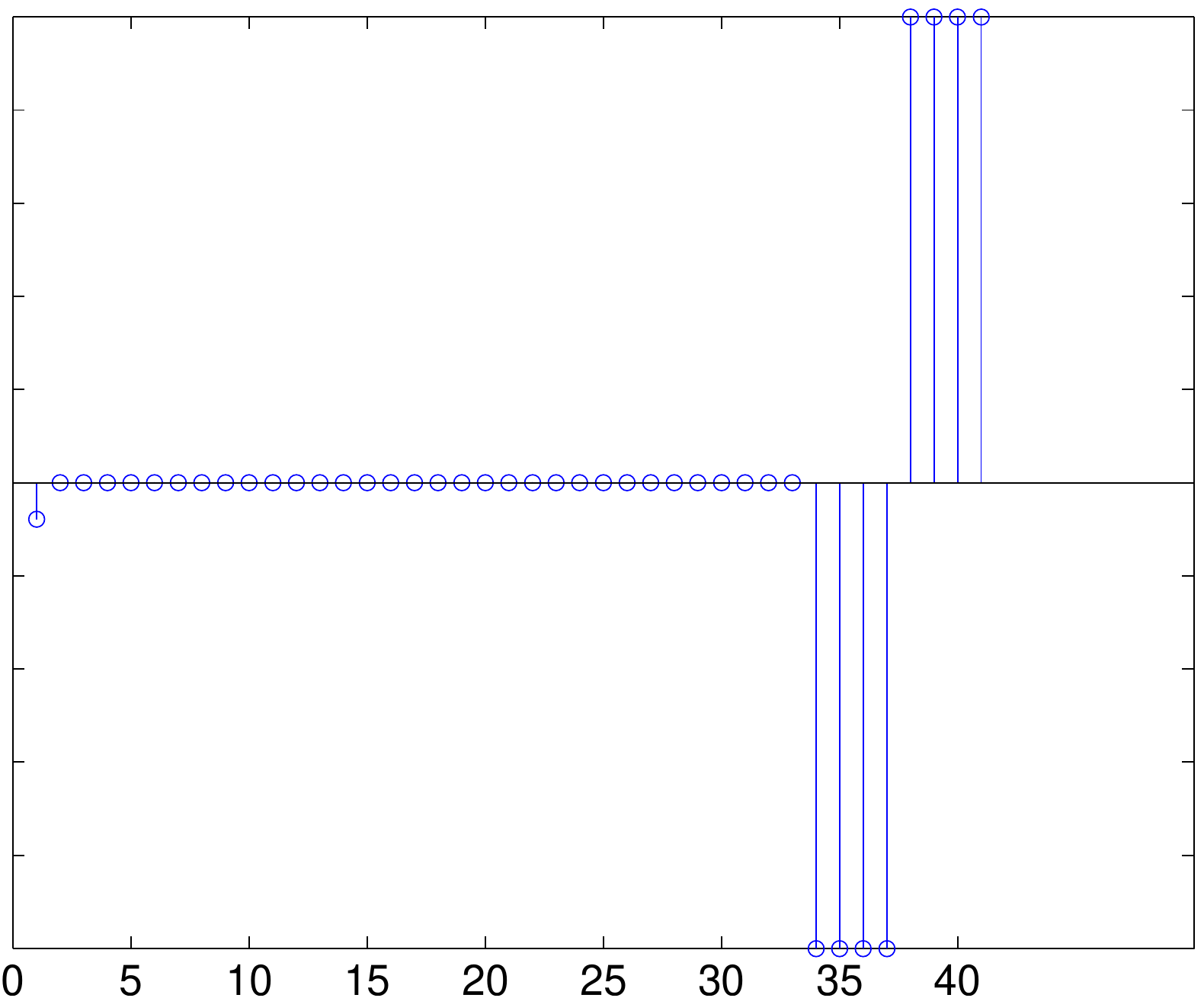}}
}
\vspace{-10pt}
\caption{Visualization of the lasso, the elastic net, the \eig and the true classifier weights. These classifiers are estimated on 80 training samples with 40 features. Among the 40 features, 8 of them (as well as the bias) are relevant for the classification task. On this dataset the test error rates of the lasso, the elastic net, and the Bayesian lasso, the \eig are 0.297, 0.245, 0.251, and 0.137.}
\label{fig:w}
\end{figure}

\subsection{Visualization of estimated classifiers}
First, we test these methods on synthetic data that contain correlated features. We sample 40 dimensional data points, each of which contains two groups of correlated variables. The correlation coefficient between variables in each group is 0.81 and there are 4 variables in each group. We set  the values of the classifier weights in one group as 5 and in the other group as -5. We also generate the bias term randomly from a standard Gaussian distribution. We set the number of training points to 80. Figure \ref{fig:w} shows the estimated classifiers and the true classifier. It is not surprising that the elastic net identifies more features than the lasso. What is interesting is that \eig does not suppress many the irrelevant features to be exactly 0, but it clearly identifies
{\em all} the relevant one, which dominate the irrelevant ones. To save space, we did not show the estimated classifier by the Bayesian lasso. Similar to the \eig, its classifier also contains many small, but nonzero weights. On this dataset, the test error rates of the lasso, the elastic net, the Bayesian lasso, and the \eig are 0.297, 0.245, 0.251, and 0.137.

\begin{figure}
\center
{
\subfigure[Bayesian lasso]{\label{fig:cov_bl}\includegraphics[scale=0.28]{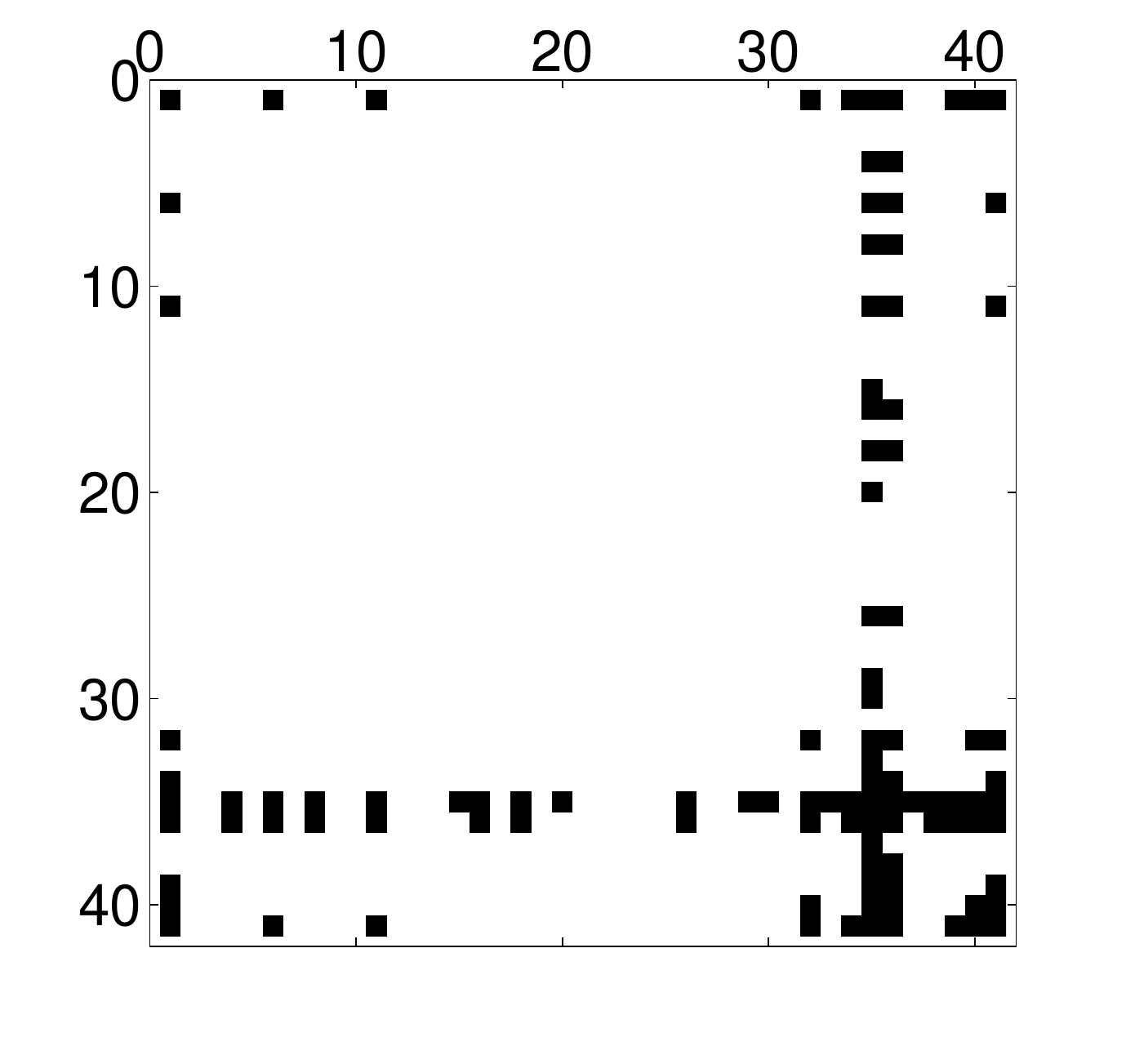}}
\subfigure[\eig]{\label{fig:cov_eig}\includegraphics[scale=0.28]{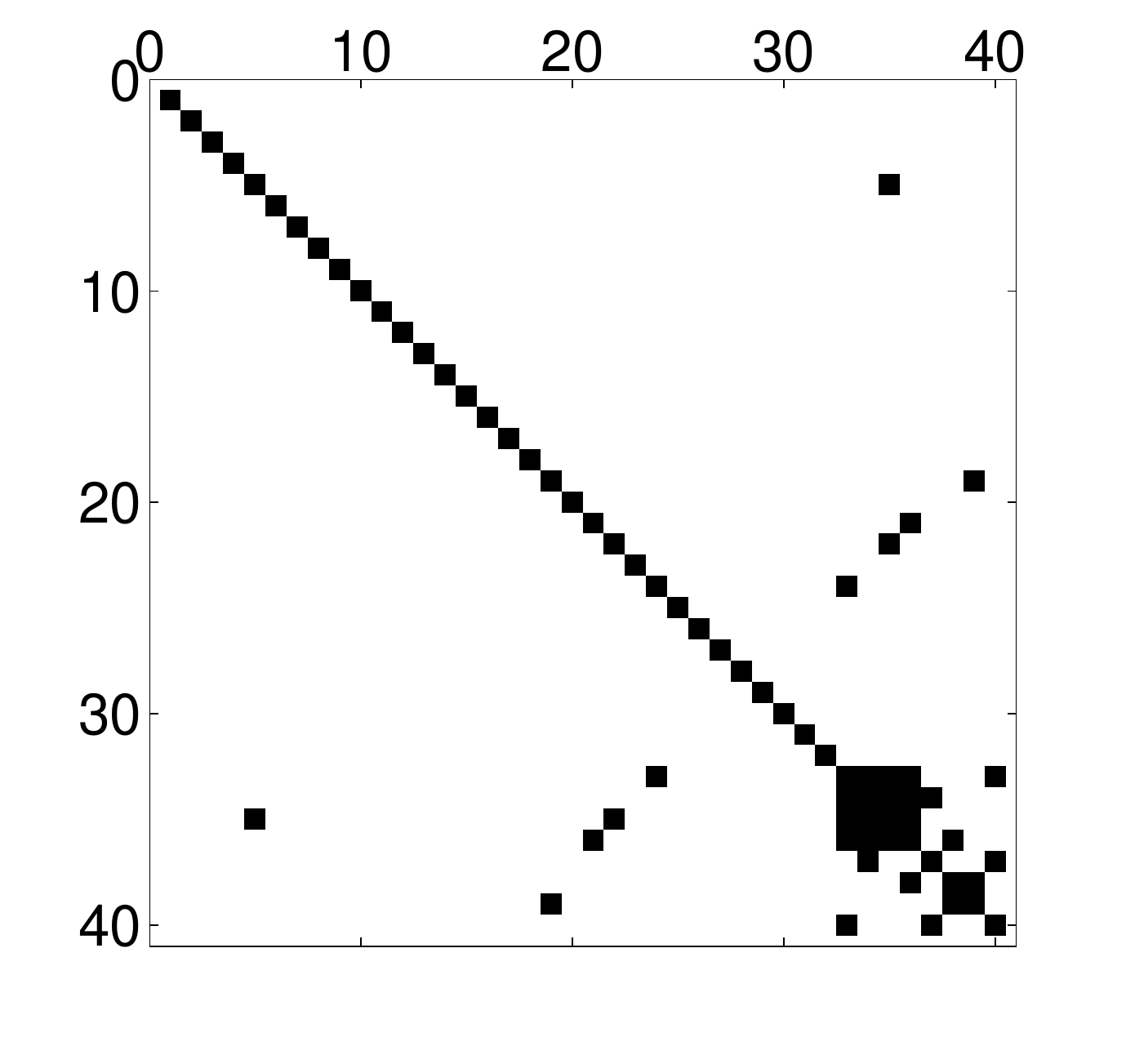}}
}
\vspace{-10pt}
\caption{Covariance matrices of the Bayesian lasso and the \eig classifiers. The covariance matrices are estimated based on the MCMC samples for these two models. We use 80 training samples with 40 features per sample. The covariance matrix of the \eig classifier correctly suggests the last few features are correlated. In particular, it clearly identifies a group of four correlated features.}
\label{fig:cov}
\end{figure}

An advantage of the Bayesian treatment for feature selection over frequentist approaches is to possibly uncover the correlations between the classifier weights. These correlations can be revealed by the covariance matrices of the joint posterior distribution over the classifier weights. In Figure \ref{fig:cov}, we visualize the quantized covariance matrices estimated by the Bayesian lasso and the \eig. As shown in \ref{fig:cov_bl} and \ref{fig:cov_eig}, while the Bayesian lasso suggests some correlation structures among features, they are fairly noisy. By contrast, the \eig  shows the two groups of correlated features  much more clearly.


\begin{figure}[h]
\center
{
\subfigure[\textit{Data with independent features}]{\label{fig:syn-ind}\includegraphics[width=2.0in]{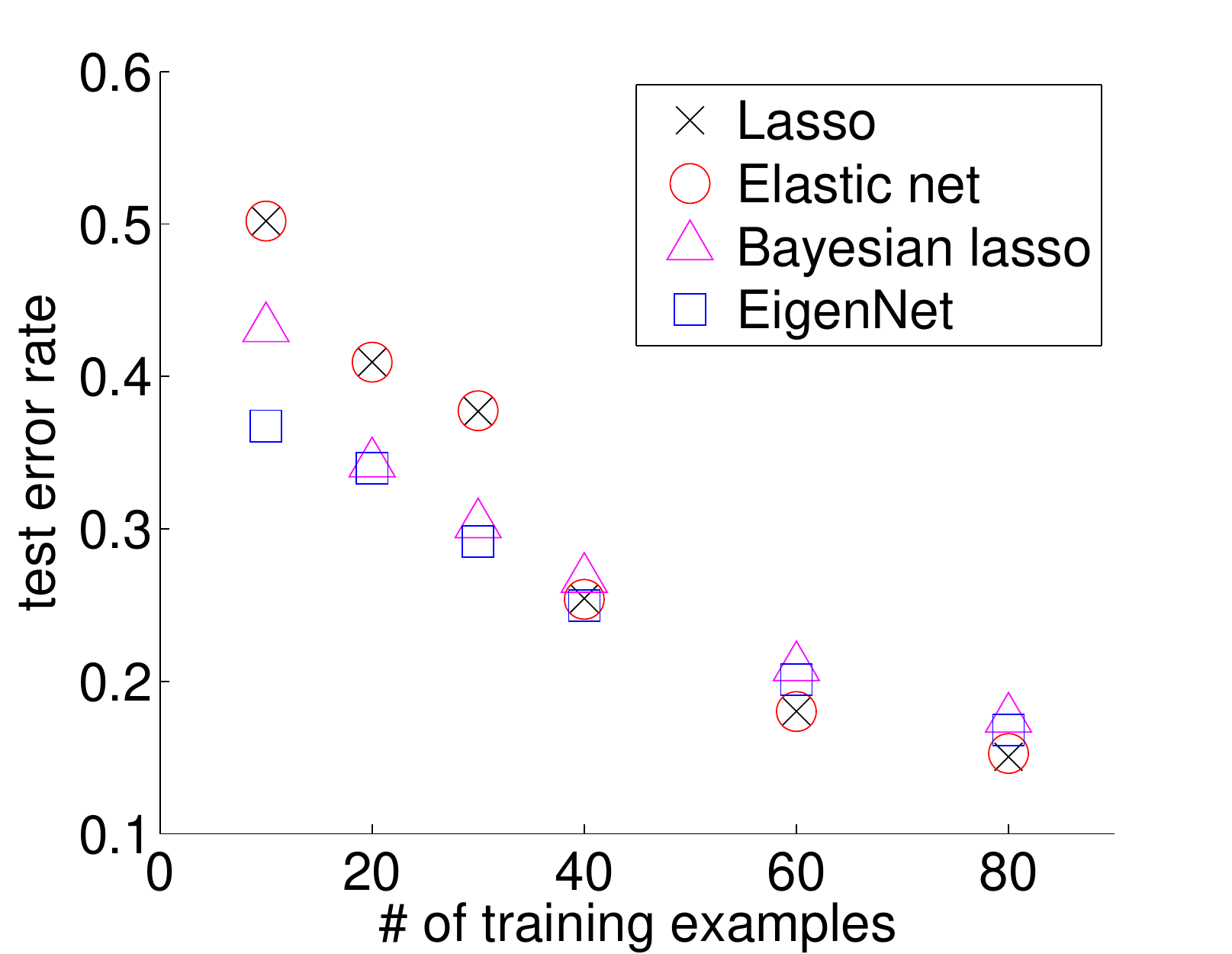}}
\hspace{.5in}
\subfigure[\textit{Data with correlated features}]{\label{fig:syn-cor}\includegraphics[width=2.0in]{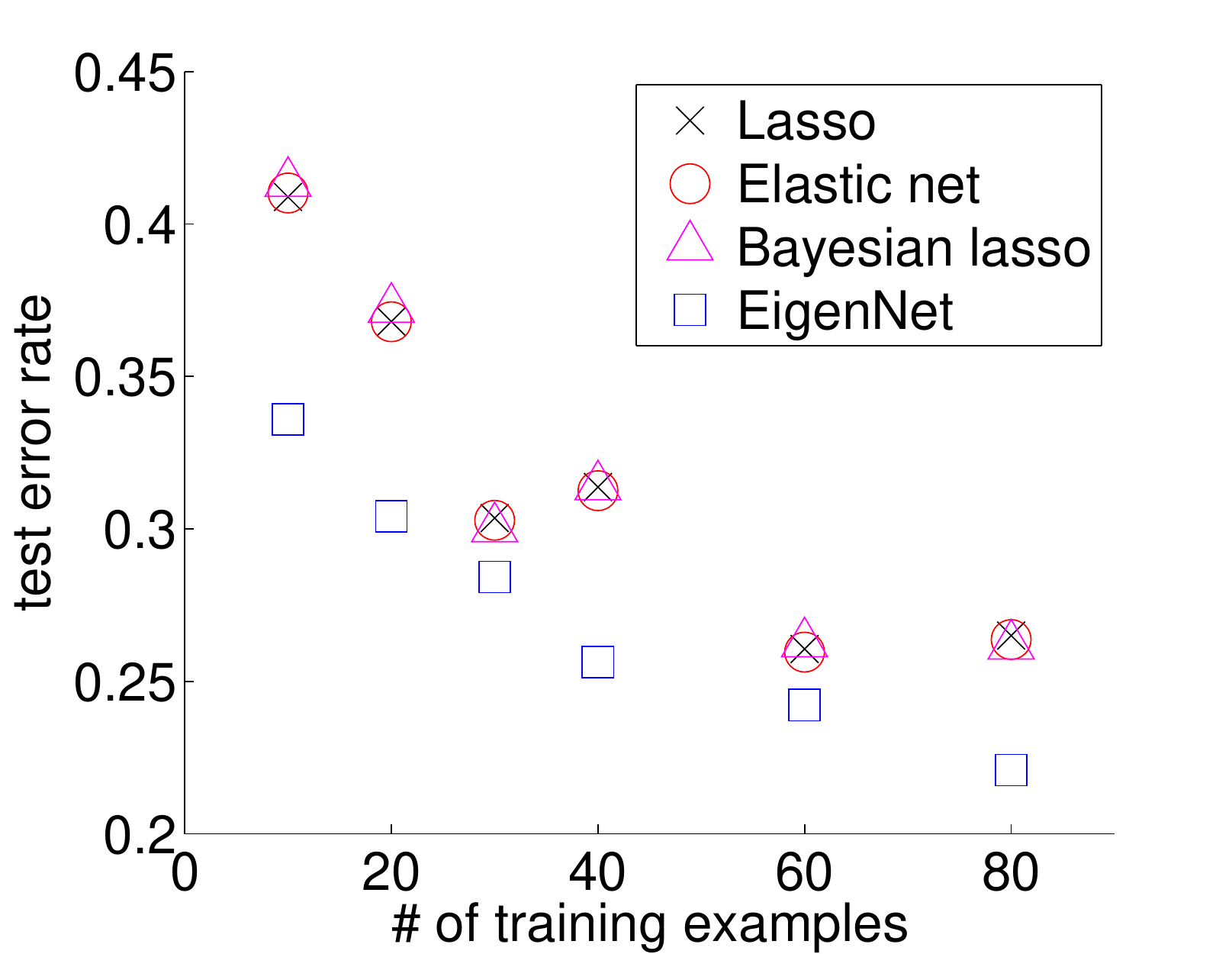}}
}
\caption{Test error rates on synthetic datasets with independent features and with correlated features.
Each training sample has 40 features, 8 of which are revelent features. We increase the number of training samples from 10 to 80 and use 2000 test samples each time. The results are averaged over 10 runs. For the data with independent features, the \eig outperforms the alternative methods at beginning when the number of training samples are fewer than 40, the number of the features. With more training samples containing independent features, all these methods perform comparably. For data with correlated features, the \eig outperforms the alternative methods consistently.}
\label{fig:syn-res}
\end{figure}

\subsection{Classification of synthetic data}
Now we systematically compare these methods on synthetic datasets containing correlated features and datasets containing independent features. For this first case, we use a similar procedure as in the visualization example: we sample 40 dimensional data points, each of which contains two groups of correlated variables. The correlation coefficient between variables in each group is 0.81 and there are 4 variables in each group. However, unlike for the previous example where the classifier weights are the same for the correlated variables, now we set the weights within the same group to have the same sign, but with different random values. We vary the number of training points, ranging from 10 to 80, and test all these methods.  For the datasets with independent features, we follow the same procedure except that the features are independently sampled.

We run the experiments 10 times. Figure \ref{fig:syn-res} shows the error rates averaged over 10 runs. 
We do not plot the standard errors of the test error rates, since they have very small values: the biggest one is less than 0.0183 for the results on data with correlated features, and for the results on data with independent features, the biggest one is less than 0.030. We report the numerical values of both the averaged error rates and the standard errors in the supplemental materials.


For the datasets with independent features, the \eig outperforms the alternative methods when the number of training samples are smaller than 40, the number of features (i.e., $p>n$). Since in this case the eigenstructures of the datasets are uninformative, we expect the improved prediction accuracy is the result of the subspace constraint used by the \eig. And once the number of training samples are not bigger than the data dimension, all these methods perform quite similarly.

For the datasets with correlated features, the \eig significantly outperforms the alternative methods consistently, not only when the number of training samples are smaller than 40 ($p>n$) but also when it is not. We believe this is because the \eig  uses the valuable eigen information revealing the feature correlations to train its classifiers. Note that although the result of the elastic net appear to overlaps with those of the lasso. Actually for the data with correlated features,  the elastic net often slightly outperforms the lasso (Please their numerical values in the supplemental materials).


\begin{figure}[h]
\center
{
\subfigure[\textit{Spambase}]{\label{fig:spambase}\includegraphics[width=2.2in]{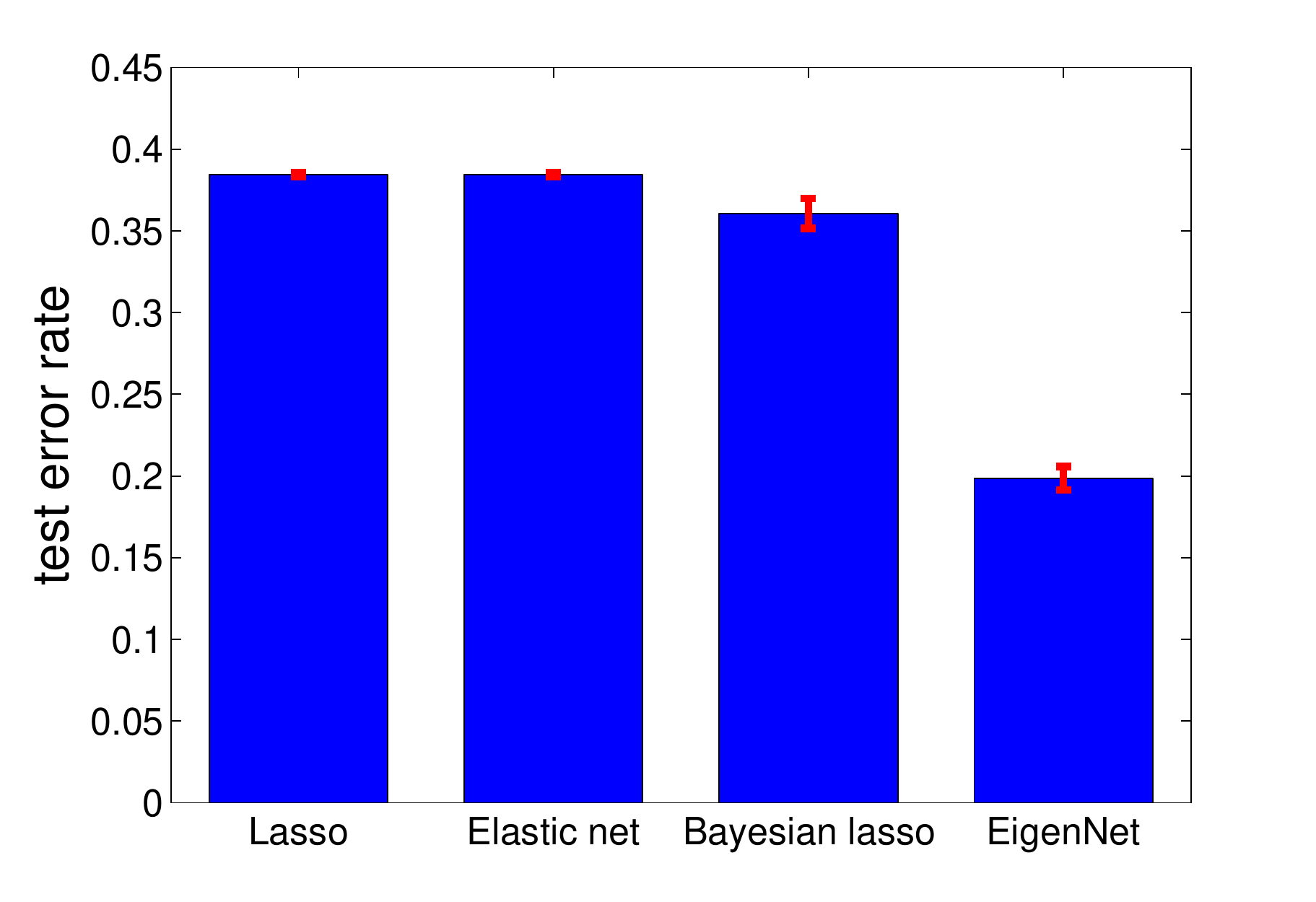}}
\subfigure[\textit{Colon}]{\label{fig:colon}\includegraphics[width=2.2in]{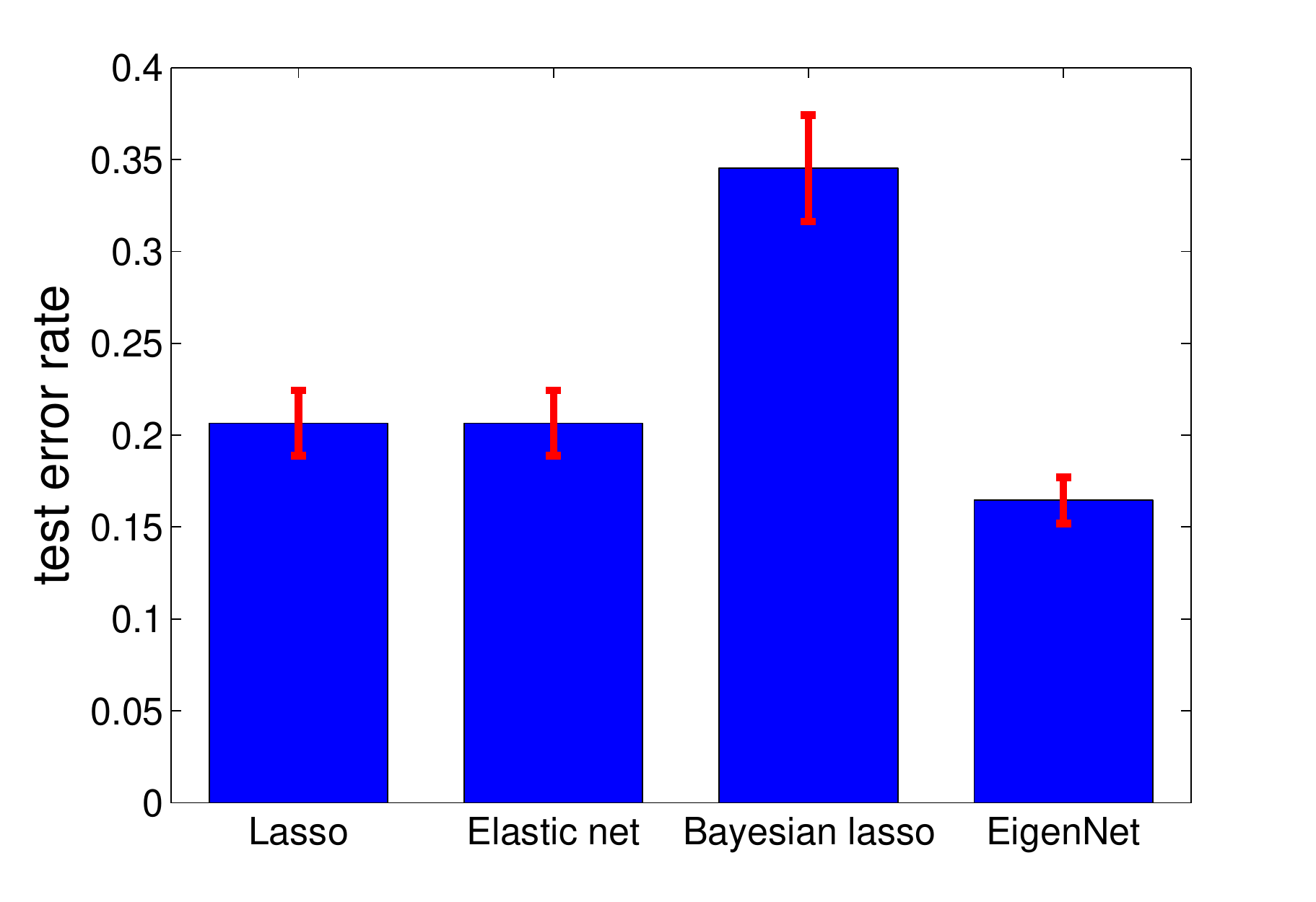}}
\subfigure[\textit{Leukemia}]{\label{fig:leuk}\includegraphics[width=2.2in]{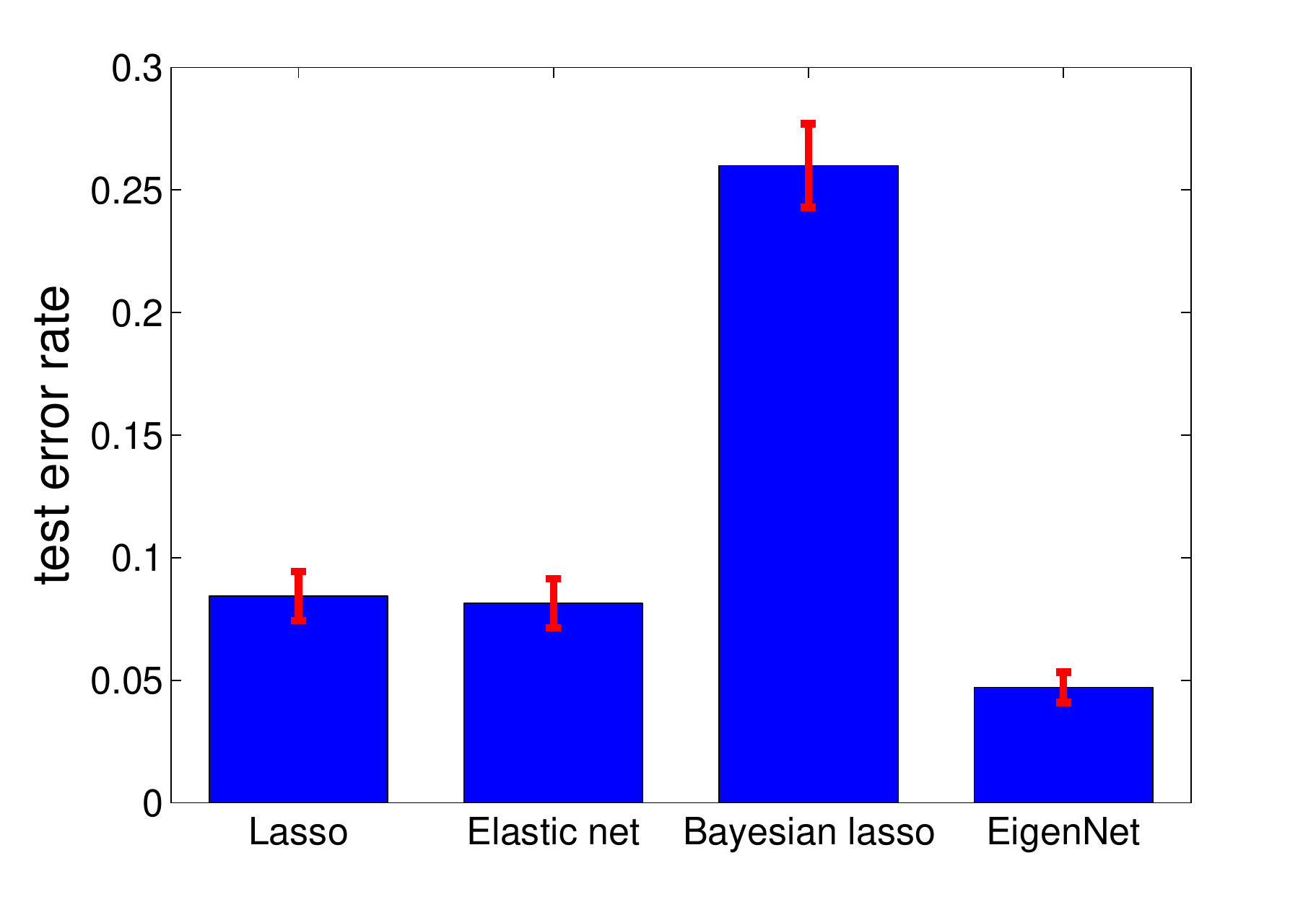}}
}
\caption{ Test error rates on spambase, leukemia and colon cancer datasets. The error bars represent the standard errors of the error rates. The results on the spambase and colon cancer datasets are averaged over 10 random partitions and the results on the leukemia dataset are averaged over 20 partitions.}
\label{fig:real-res}
\end{figure}


\subsection{Classification of real data}
Besides the synthetic data, we also test all these methods on UCI benchmark datasets,
two high-dimensional gene expression datasets, leukaemia and colon cancer, and a spambase dataset with relatively lower dimension but a lot more training samples.

For the leukaemia dataset, the task is to distinguish acute myeloid leukaemia (AML) from acute lymphoblastic
leukaemia (ALL). The whole dataset has 47 and 25 samples of type ALL and AML respectively with
7129 features per sample. The dataset was randomly split 20 times into 37 training and 35 test samples.

For the colon cancer dataset, the task is to discriminate tumor from normal tissues using microarray
data. The dataset has 22 normal and 40 cancer samples with 2000 features per sample. We randomly
split the dataset into 31 training and 31 test samples 10 times.

For the spambase datast, the task is to detect spam emails, i.e., unsolicited commercial emails.
We use 57 features indicating whether a particular word or character was frequently occurring in the emails. We randomly
split the dataset into 1533 training and 3066 test samples 10 times. Note that we do not use any kernel here and the results on this dataset
are meant to examine how the performance of these methods compares to each other when there are more samples than features. Using a nonlinear basis function, e.g., a radial basis function, 
is expected to boost the predictive performance of all these methods.

Figure \ref{fig:real-res} summarizes the average test error rates and the standard errors of these methods on the three datasets. Again, the \eig significantly outperforms the alternative methods on three datasets. Note that for the leukaemia and colon cancer datasets Bayesian lasso does not perform much worse than the other methods. The reason, we believe, is that these two high dimensional datasets contain thousands of features and Bayesian lasso directly draws samples in such high dimensional spaces, leading to very slow mixing rates. By contrast, the \eig draws samples efficiently in a much smaller eigenspace, not only leading to faster mixing rates but also greatly saving the computing cost for obtaining each sample.


\section{Conclusions}
In this paper, we have presented a novel sparse Bayesian hybrid model,
the \eig. It integrates a sparse conditional classification model with a generative model capturing the feature correlations. It also generalizes the elastic net by explicitly exploring correlations between features. Compared with several state-of-the art methods, the \eig achieves significantly improved prediction accuracy on several benchmark datasets.

We plan to extend our hybrid model by utilizing other probabilistic generative models, such as sparse principle component analysis and related projection methods \cite{GuanDy09,ArcBach09} and independent component analysis models. Compared to the classical PCA models, these models could be used to better guide the selection of interdependent sparse features.

\subsection*{Acknowledgement}

Thanks to Jyotishka Datta for his help on software implementation and to Tommi Jaakkola for stimulating discussion.


\bibliography{EigenNet_ref}
\bibliographystyle{unsrtnat}

\appendix
{\noindent{\bf Appendix}}

Given the linear relationship between $(\balpha, \bbeta)$ and $(\w, \tw)$, the prior $p(\w, \tw)$ defined in \eqref{eq:prior_joint} is equivalent to  $p(\balpha, \bbeta)$ defined in \eqref{eq:prior0}.

First, when $n\geq p$, we can easily obtain the $p(\balpha, \bbeta)$ from $p(\w, \tw)$. In this case, the number of eigenvectors is $p$ and the Jacobian matrix is the $p \times p$ full rank matrix $\V$. Furthermore, the determinant of $\V$ is 1 since $\V$ is an orthonormal matrix. Therefore, with $[\w,  \tw]= \V [\balpha, \bbeta]$ we have $p(\balpha, \bbeta)$ = $p(\w, \tw)$.

When $p>n$, $\V_{p\times n}$ is a tall matrix and therefore we cannot compute its determinant to transform the prior distribution $p(\balpha, \bbeta)$. Now $p(\w,\tw)$ is essentially a distribution on the data subspace embedded in the high dimensional space ${\cal R}^p$. To obtain the equivalence between these two priors, we consider the following theorem \cite{Petersen08cookbook}:
        \begin{theorem}
            If $\A$ is ``tall", i.e.,``under-determined", then
            $
            p(\x)=\int p(\s)\delta(\x-\A\s)\dif \s =
            \begin{cases}
            \frac{1}{\sqrt{|\A^T \A|}}p(\A^+ \x) \quad if \quad \x=\A\A^+\x \\
            \0 \quad\textrm{ otherwise}
            \end{cases}
            $
         \end{theorem}
Using this theorem and the fact $|\V\V^+|=1$, we see that with the simple linear relationship between the variables, $p(\balpha, \bbeta)$ = $p(\w, \tw)$.

\end{document}